%% file: egpaper_for_review.tex
\documentclass[10pt,twocolumn,letterpaper]{article}

\usepackage{cvpr}
\usepackage{times}
\usepackage{epsfig}
\usepackage{graphicx}
\usepackage{amsmath}
\usepackage{amssymb}

\usepackage{algpseudocode}
\usepackage{algorithm}
\usepackage{array}
\usepackage{multirow}
\usepackage{dsfont}

\usepackage[pagebackref=true,breaklinks=true,letterpaper=true,colorlinks,bookmarks=false]{hyperref}

\cvprfinalcopy 

\def\cvprPaperID{5904} 

\ifcvprfinal\fi

\begin{document}

\title{Learning from Synthetic Animals}

\author{Jiteng Mu\thanks{$^∗$ Indicates equal contributions.}, Weichao Qiu\footnotemark[1], Gregory Hager, Alan Yuille\\
Johns Hopkins University\\
{\tt\small jitengmu@jhu.edu, hager@cs.jhu.edu, \{qiuwch, alan.l.yuille\}@gmail.com}
}

\maketitle

\begin{abstract}

Despite great success in human parsing, progress for parsing other deformable articulated objects, like animals, is still limited by the lack of labeled data. In this paper, we use synthetic images and ground truth generated from CAD animal models to address this challenge. To bridge the domain gap between real and synthetic images, we propose a novel consistency-constrained semi-supervised learning method (CC-SSL). Our method leverages both spatial and temporal consistencies, to bootstrap weak models trained on synthetic data with unlabeled real images. We demonstrate the effectiveness of our method on highly deformable animals, such as horses and tigers. Without using any real image label, our method allows for accurate keypoint prediction on real images. Moreover, we quantitatively show that models using synthetic data achieve better generalization performance than models trained on real images across different domains in the Visual Domain Adaptation Challenge dataset. Our synthetic dataset contains 10+ animals with diverse poses and rich ground truth, which enables us to use the multi-task learning strategy to further boost models' performance.

\end{abstract}

\input{01_intro.tex}
\input{02_related.tex}
\input{03_method.tex}
\input{04_exp.tex}

\section{Conclusions}
In this paper, we present a simple yet efficient method using synthetic images to parse animals. To bridge the domain gap, we present a novel consistency-constrained semi-supervised learning (CC-SSL) method, which leverages both spatial and temporal constraints. We demonstrate the effectiveness of the proposed method on horses and tigers in the TigDog Dataset. Without any real image label, our model can detect keypoints reliably on real images. When using real image labels, we show that models trained jointly on synthetic and real images achieve better results compared to models trained only on real images. We further demonstrate that the models trained using synthetic data achieve better generalization performance across different domains in the Visual Domain Adaptation Challenge dataset. We build a synthetic dataset contains 10+ animals with diverse poses and rich ground truth and show that multi-task learning is effective.

\section*{Acknowledgements}
\vspace{-0.2cm}
Supported by the Intelligence Advanced Research Projects Activity (IARPA) via Department of Interior/Interior Business Center (DOI/IBC) contract number D17PC00342. The U.S. Government is authorized to reproduce and distribute reprints for Governmental purposes notwithstanding any copyright annotation thereon. Disclaimer: The views and conclusions contained herein are those of the authors and should not be interpreted as necessarily representing the official policies or endorsements, either expressed or implied, of IARPA, DOI/IBC, or the U.S. Government. The authors would like to thank Chunyu Wang, Qingfu Wan, Yi Zhang for helpful discussions.

{\small
\bibliographystyle{ieee_fullname}
\bibliography{egbib}
}

\end{document}

%% file: 01_intro.tex
\section{Introduction}

Thanks to the presence of large scale annotated datasets and powerful Convolutional Neural Networks(CNNs), the state of human parsing has advanced rapidly. By contrast, there is little previous work on parsing animals. Parsing animals is important for many tasks, including, but not limited to monitoring wild animal behaviors, developing bio-inspired robots, building motion capture systems, etc. 

One main problem for parsing animals is the limit of datasets. Though many datasets containing animals are built for classification, bounding box detection, and instance segmentation, only a small number of datasets are built for parsing animal keypoints and parts. Annotating large scale datasets for animals is prohibitively expensive. Therefore, most existing approaches for parsing humans, which often require enormous annotated data  \cite{DBLP:conf/cvpr/AndrilukaPGS14,DBLP:conf/cvpr/SappT13}, are less suited for parsing animals. 

In this work, we use synthetic data to address this challenge. Many works \cite{DBLP:conf/cvpr/TremblayPABJATC18,DBLP:conf/icra/PrakashBBACSSB19} show that by jointly using synthetic images and real images, models can yield supreme results. In addition, synthetic data also has many unique advantages compared to real-world datasets. First, rendering synthetic data with rich ground truth at scale is easier and cheaper compared with capturing and annotating real-world images. Second, synthetic data can also provide accurate ground truth for cases where annotations are hard to acquire for natural images, such as labeling optical flow \cite{DBLP:conf/iccv/DosovitskiyFIHH15} or under occlusion and low-resolution. Third, real-world datasets usually suffer from the long-tail problem where rare cases are less represented. Generated synthetic datasets can avoid this problem by sampling rendering parameters. 


\begin{figure*}[h]
    \centering
    \includegraphics[width=\linewidth,height=6.5cm]{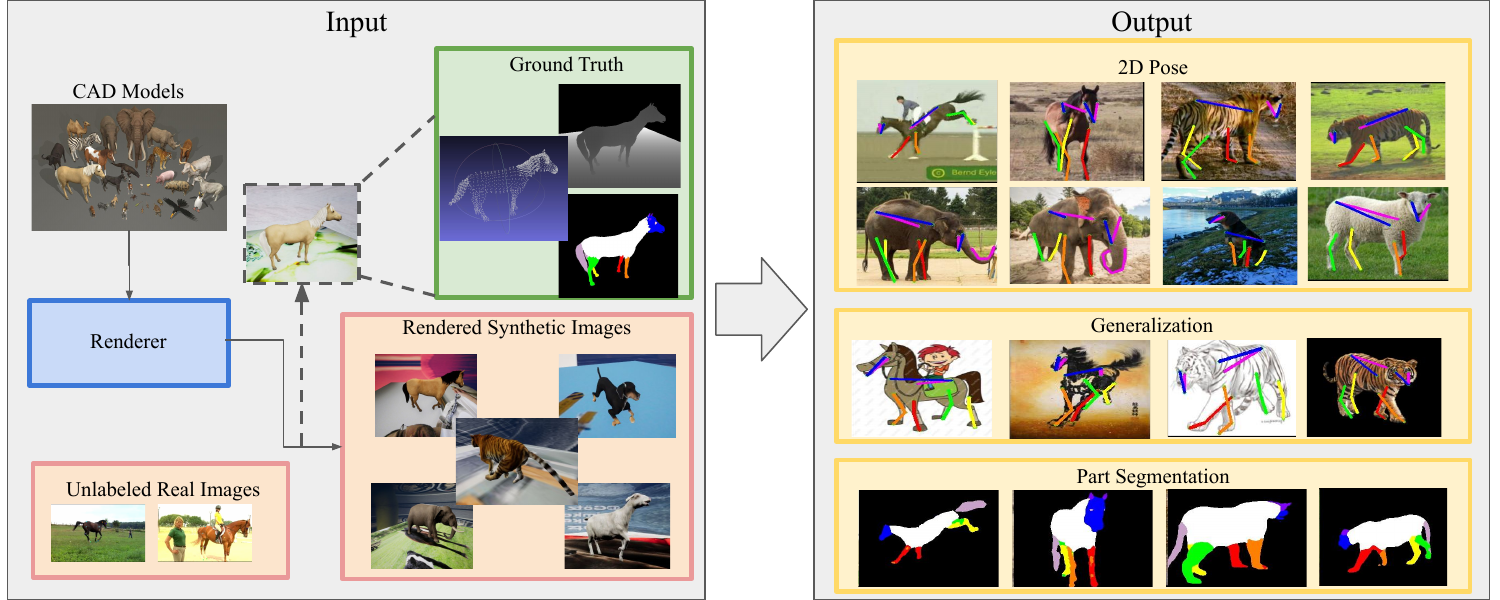}
    \caption{Overview. We generate a synthetic animal dataset by randomly sampling rendering parameters including camera viewpoints, lighting, textures and poses. The dataset contains 10+ animals along with rich ground truth, such as dense 2D keypoints, part segmentation and depth maps. With the synthetic dataset, we propose an effective method which allows for accurate keypoint prediction across domains. In addition to 2D pose estimation, we also show models can predict accurate part segmentation.}
    \label{fig:overview}
\end{figure*}

However, there are large domain gaps \cite{DBLP:conf/3dim/ChenWLSWTLCC16,DBLP:conf/cvpr/Varol0MMBLS17,DBLP:conf/icml/HoffmanTPZISED18} between synthetic images and real images, which prevent models trained on synthetic data from generalizing well to real-world images. Moreover, synthetic data is also limited by object diversity. ShapeNet \cite{DBLP:journals/corr/ChangFGHHLSSSSX15} has been created to include diverse 3D models and SMPL \cite{DBLP:journals/tog/LoperM0PB15} has been built for humans. Nevertheless, creating such diverse synthetic models is a difficult task, which requires capturing the appearance and attaching a skeleton to the object. Besides, considering the number of animal categories in the world, creating diverse synthetic models along with realistic textures for each animal is almost infeasible.

In this paper, we propose a method where models are trained using synthetic CAD models. Our method can achieve high performance with only a single CAD animal model. We generate pseudo-labels on unlabeled real images for semi-supervised learning. To handle noisy pseudo-labels, we design three consistency-check criteria to evaluate the quality of the predicted labels, which we refer to as consistency-constrained semi-supervised learning (CC-SSL). Through extensive experiments, we show that our models achieve similar performance to models trained on real data, but without using any annotation of real images. It also outperforms other domain adaptation methods by a large margin. Providing real image annotations, the performance can be further improved. Furthermore, we demonstrate models trained with synthetic data show better domain generalization performance in multiple visual domains compared with those trained on real data. The code is available at \url{https://github.com/JitengMu/Learning-from-Synthetic-Animals}.

We summarize the contributions of our paper as follows. 
First, we propose a consistency-constrained semi-supervised learning framework (CC-SSL) to learn a model with one single CAD object. We show that models trained with synthetic data and unlabeled real images allow for accurate keypoint prediction on real images.
Second, when using real image labels, we show that models trained jointly on synthetic and real images achieve better results compared to models trained only on real images.
Third, we evaluate the generalizability of our learned models across different visual domains in the Visual Domain Adaptation Challenge dataset and we quantitatively demonstrate that models trained using synthetic data show better generalization performance than models trained on real images.
Lastly, we generate an animal dataset with 10+ different animal CAD models and we demonstrate the data can be effectively used for 2D pose estimation, part segmentation, and multi-task learning.

%% file: 02_related.tex
\section{Related Work}

    \subsection{Animal Parsing}
Though there exists large scale datasets containing animals for classification, detection, and instance segmentation, there are only a small number of datasets built for pose estimation \cite{DBLP:conf/cvpr/PeroRSF15,WelinderEtal2010,DBLP:journals/corr/abs-1908-05806,DBLP:conf/bmvc/NovotnyLV16,DBLP:journals/corr/abs-1906-05586} and animal part segmentation \cite{DBLP:conf/cvpr/ChenMLFUY14}. Besides, annotating keypoints or parts is time-consuming and these datasets only cover a tiny portion of animal species in the world. 

Due to the lack of annotations, synthetic data has been widely used to address the problem \cite{DBLP:journals/corr/abs-1908-07201,DBLP:journals/corr/abs-1811-05804,DBLP:conf/cvpr/ZuffiKB18,DBLP:conf/cvpr/ZuffiKJB17}. Similar to SMPL models \cite{DBLP:journals/tog/LoperM0PB15} for humans, \cite{DBLP:conf/cvpr/ZuffiKJB17} proposes a method to learn articulated SMAL shape models for animals. Later, \cite{DBLP:conf/cvpr/ZuffiKB18} extracts more 3D shape details and is able to model new species. Unfortunately, these methods are built on manually extracted silhouettes and keypoint annotations. Recently, \cite{DBLP:journals/corr/abs-1908-07201} proposes to copy texture from real animals and predicts 3D mesh of animals in an end-to-end manner. Most related to our method is \cite{DBLP:journals/corr/abs-1811-05804}, where authors propose a method to estimate animal poses on real images using synthetic silhouettes. Different from \cite{DBLP:journals/corr/abs-1811-05804} which requires an additional robust segmentation model for real images during inference, our strategy does not require any additional models.
    
    \subsection{Unsupervised Domain Adaptation}
Unsupervised domain adaptation focuses on learning a model that works well on a target domain when provided with labeled source samples and unlabeled target samples. A number of image-to-image translation methods \cite{DBLP:conf/nips/LiuBK17,DBLP:conf/iccv/ZhuPIE17,DBLP:conf/eccv/HuangLBK18} are proposed to transfer images from different domains. Another line of work studies how to explicitly minimize some measure of feature difference, such as maximum mean discrepancy \cite{DBLP:journals/corr/TzengHZSD14,DBLP:conf/icml/LongC0J15} or correlation distances \cite{DBLP:conf/eccv/SunS16,DBLP:conf/iccv/TzengHDS15}. \cite{DBLP:conf/nips/BousmalisTSKE16} proposes to explicitly partition features into a shared space and a private space. Recently, adversarial loss \cite{DBLP:conf/cvpr/TzengHSD17,DBLP:conf/icml/HoffmanTPZISED18} is used to learn  domain invariant features, where a domain classifier is trained to distinguish the source and target distributions. \cite{DBLP:conf/cvpr/TzengHSD17} proposes a general framework to bring features from different domains closer. \cite{DBLP:conf/icml/HoffmanTPZISED18,DBLP:conf/cvpr/MurezKKRK18} extend this idea with cycle consistency to improve results. 

Recent works have also investigated how to use these techniques to advance deformable objects parsing. \cite{DBLP:conf/3dim/ChenWLSWTLCC16} studies using synthetic human images combined with domain adaptation to improve human 3D pose estimation. \cite{DBLP:conf/cvpr/Varol0MMBLS17} renders 145 realistic synthetic human models to reduce the domain gap. Different from previous works where a large amount of realistic synthetic models are required, we show that models trained on one CAD model can learn domain-invariant features.

    \subsection{Self-training}
Self-training has been proved effective in semi-supervised learning. Early work \cite{lee2013pseudo} draws the connection between deep self-training and entropy regularization. However, since generated pseudo-labels are noisy, a number of methods \cite{DBLP:journals/corr/abs-1908-07387,DBLP:conf/wacv/DingWFG18,DBLP:conf/eccv/ZouYKW18,DBLP:journals/corr/abs-1908-09822,DBLP:conf/iclr/LaineA17,DBLP:conf/iclr/FrenchMF18,DBLP:conf/cvpr/LiYV19,DBLP:journals/corr/abs-1909-00589,DBLP:conf/cvpr/RadosavovicDGGH18,DBLP:conf/cvpr/RoyChowdhuryCSJ19} are proposed to address the problem. \cite{DBLP:conf/eccv/ZouYKW18,DBLP:journals/corr/abs-1908-09822} formulate self-training as a general EM algorithm and proposes a confidence regularized self-training framework. \cite{DBLP:conf/iclr/LaineA17} proposes a self-ensembling framework to bootstrap models using unlabeled data. \cite{DBLP:conf/iclr/FrenchMF18} extends the previous work to unsupervised domain adaptation and demonstrate its effectiveness in bridging domain gaps. 

Closely related to our work on 2D pose estimation is \cite{DBLP:conf/cvpr/RadosavovicDGGH18}, where the authors propose a simple method for omni-supervised learning that distills knowledge from unlabeled data and demonstrate its effectiveness on detection and pose estimation. However, under large domain discrepancy, the assumption that the teacher model assigns high-confidence pseudo-labels is not guaranteed. To tackle the problem, we introduce a curriculum learning strategy \cite{DBLP:conf/icml/BengioLCW09,DBLP:conf/eccv/GuoHZZDSH18,DBLP:conf/aaai/JiangMZSH15} to progressively increase pseudo-labels and train models in iterations. We also extend \cite{DBLP:conf/cvpr/RadosavovicDGGH18} by leveraging both spatial and temporal consistencies.

%% file: 03_method.tex
\section{Approach}

\begin{figure*}[h]
    \centering
    \includegraphics[width=\linewidth,height=6.5cm]{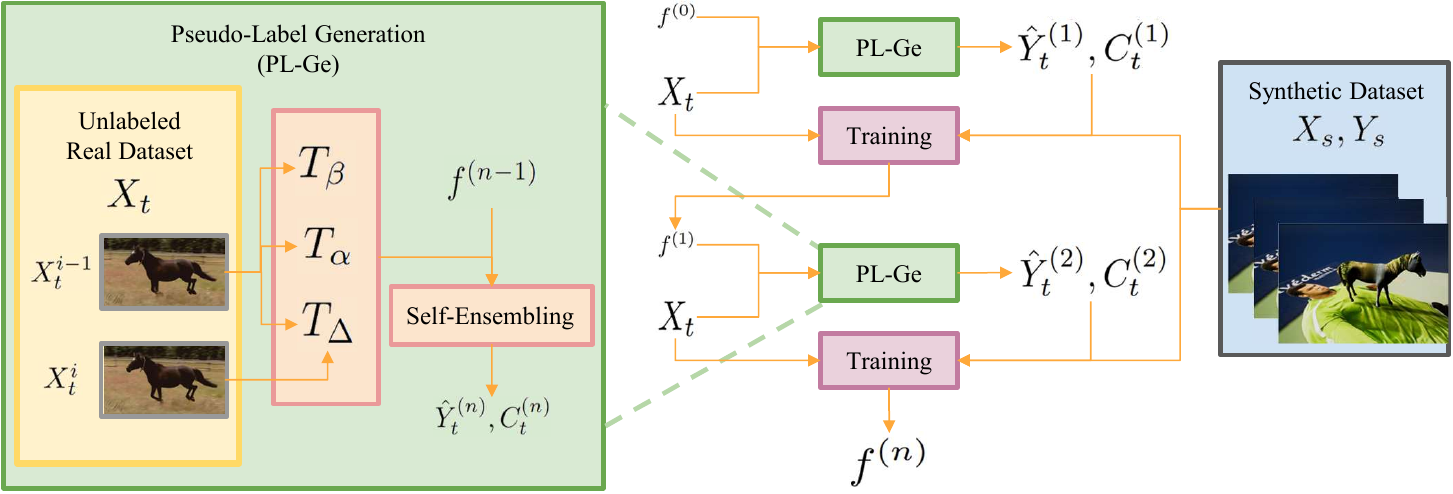}
    \caption{Consistency-constrained semi-supervised learning pipeline. $T_{\beta}$ indicates the invariance consistency, $T_{\alpha}$ indicates the equivariance consistency and $T_{\Delta}$ indicates the temporal consistency. The training procedure can be described as follows: we start with training a model only using synthetic data and obtain an initial model $f^{(0)}$. Then we iterate the following procedure. For the $n$th iteration, we first use the proposed Pseudo-Label Generation Algorithm \ref{alg} to generate labels $\hat{Y}_{t}^{(n)}$. Next, we  train the model using ($X_s, Y_s$) and ($X_t, \hat{Y}_{t}^{(n)}$) jointly.}
    \label{fig:pipeline}
\end{figure*}

    We first formulate a unified image generation procedure in Section \ref{Formulate Image Generation Procedure} built on the low dimension manifold assumption. In Section \ref{Consistency}, we define three consistencies and discuss how to take advantage of these consistencies during pseudo-label generation process. In Section \ref{Pseudo-Label Generation}, we propose a Pseudo-Label Generation algorithm using consistency-check. Then in Section \ref{Joint Training} we present our consistency-constrained semi-supervised learning algorithm and discuss the iterative training pipeline. Lastly, in Section \ref{Synthetic Dataset Generation}, we explain how our synthetic datasets are generated.
    
    We consider the problem under unsupervised domain adaptation framework with two datasets. We name our synthetic dataset as the source dataset ($X_s, Y_s$) and real images as the target dataset $X_t$. The goal is to learn a model $f$ to predict labels for the target data $X_t$. We simply start with learning a source model $f_s$ using paired data ($X_s, Y_s$) in a fully supervised way. Then we bootstrap the source model using target dataset with consistency-constrained semi-supervised learning. An overview of the pipeline is presented in Figure~\ref{fig:pipeline}. 

\subsection{Formulate Image Generation Procedure}\label{Formulate Image Generation Procedure}
    
    In order to learn a model using synthetic data that can generalize well to real data, one needs to assume that there exists some essential knowledge shared between these two domains. Take animal 2D pose estimation as an example, though synthetic and natural images look differently by textures and background, they are quite similar in terms of poses and shape. Actually, these are exactly what we hope a model trained on synthetic data can learn. So an ideal model should be able to capture these essential factors and ignore those less relevant ones, such as lighting and background.
    
    
    Formally, we introduce a generator $G$ that transforms poses, shapes, viewpoints, textures, etc, into an image. Mathematically, we group all these factors into two categories, task-related factors $\alpha$, which is what a model cares about, and others $\beta$, which are irrelevant to the task at hand. So we parametrize the image generation process as follows,
\begin{align}\label{eq:generation-model}
    X = G(\alpha,\beta)
\end{align}
    where $X$ is a generated image and $G$ denotes the generator. Specifically, for 2D pose estimation, $\alpha$ represents factors related to the 2D keypoints, such as pose and shape; $\beta$ indicates factors independent of $\alpha$, which could be textures, lighting and background.

\subsection{Consistency}\label{Consistency}

    Based on the formulation in Section \ref{Formulate Image Generation Procedure}, we define three consistencies and discuss how to take advantage of these consistencies for the pseudo-label generation process.
    
    Since model-generated labels on the target dataset are noisy, one needs to tell the model which predictions are correct and which are wrong. Intuitively, an ideal 2D keypoint detector should generate consistent predictions on one image no matter how the background is perturbed. In addition, if one rotates the image, the prediction should change accordingly as well. Based on these intuitions, we propose to use consistency-check to reduce false positives.
    
    In the following paragraphs, we will introduce invariance consistency, equivariance consistency and temporal consistency. We will discuss how to use consistency-check to generate pseudo-labels, which serves as the basis for the proposed semi-supervised learning method.
    
    

    The transformation applied to an image can be considered as directly transforming the underlying factors in Equation~\ref{eq:generation-model}. We define a general tensor operator, $T:\mathbb{R}^{H\times W}\rightarrow\mathbb{R}^{H\times W}$. In addition, we introduce $\tau_{\alpha}$ corresponding to operations that would affect $\alpha$ and $\tau_{\beta}$ to represent operations independent of $\alpha$. Then Equation~\ref{eq:generation-model} can be expressed as following,
\begin{align}\label{eq:generation-factor}
    T(X) = G(\tau_{\alpha}(\alpha),\tau_{\beta}(\beta))
\end{align}

     We use $f:\mathbb{R}^{H\times W}\rightarrow\mathbb{R}^{H\times W}$ to denote a perfect 2D pose estimation model. When $f$ is applied to Equation~ \ref{eq:generation-factor}, it is obvious that, $f[T(X)] = f[G(\tau_{\alpha}(\alpha),\tau_{\beta}(\beta))]$.

    \textbf{Invariance consistency}: If the transform $T$ does not change factors associated with the task, the model's prediction is expected to be the same. The idea here is that a well-behaved model should be invariant to operations on $\beta$. For example, in 2D pose estimation, adding noise to the image or perturbing colors should not affect the model's prediction. We name these transforms \textit{invariance transform $T_\beta$}, as shown in Equation~\ref{eq:consistency1}.
\begin{align}\label{eq:consistency1}
    f[T_{\beta}(X)] = f(X)
\end{align}
    
    If we apply multiple invariance transforms to the same image, the predictions on these transformed images should be consistent. This consistency can be used to verify whether the prediction is correct, which we refer to as \textit{invariance consistency}. 

    \textbf{Equivariance consistency}: Besides invariance transform, there are other cases where the task related factors are changed. We use $T_{\alpha}$ to denote transforms related to operations $\tau_{\alpha}$. There are special cases where we can easily get the corresponding $T_{\alpha}$. One easy case is that, sometimes, the effect of $\tau_{\alpha}$ only cause geometric transformations in 2D images, which we refer to as \textit{equivariance transform $T_\alpha$}. Actually, this is essentially similar to what \cite{DBLP:conf/cvpr/RadosavovicDGGH18} proposes. Therefore, we have \textit{equivariance consistency} as shown in Equation~\ref{eq:consistency2}.
\begin{align}\label{eq:consistency2}
\begin{split}
    f[T_{\alpha}(X)] = T_{\alpha}[f(X)]
\end{split}
\end{align}
    It is also easy to show that $f(X) = T_{\alpha}^{-1}[f[T_{\alpha}(X)]]$, which means that, after applying the inverse transform $T_{\alpha}^{-1}$, a good model should give back the original prediction. 

    \textbf{Temporal consistency}: It is difficult to model transformations between frames in a video. This transform $T_{\Delta}$ does not satisfy the invariance and equivariance properties described above. However, $T_{\Delta}$ is still induced by variations of underlying factors $\alpha$ and $\beta$. It is reasonable to assume that, in a real-world video, these factors do not change dramatically between neighboring frames.
\begin{align}\label{eq:consistency3}
\begin{split}
    f[T_{\Delta}(X)] = f(X) + \Delta
\end{split}
\end{align}
    So we assume the keypoints shifting between two frames is relatively small as shown in Equation~\ref{eq:consistency3}. Intuitively, this means that the keypoint prediction for the same joint in consecutive frames should not be too far away, otherwise it is likely to be incorrect.
    
    For 2D keypoint estimation, we observe that $T_{\Delta}$ can be approximated by  optical flow to get $\Delta$, which allows us to use optical flow to propagate pseudo-labels from confident frames to less confident ones.
    


    Although we define these three consistencies for  2D pose estimation, they can be easily extended to other problems. For example, in 3D pose estimation, $\alpha$ can be factors related to 3D pose. Then the invariance consistency is still the same, but the equivariance consistency no longer holds, since the mapping of 3D pose to 2D pose is not a one-to-one mapping and there are ambiguities in the depth dimension. However, one can still use it as a constraint for the other two dimensions, which means the projected poses should still satisfy the same consistency. So it is easy to see that though corresponding consistencies may  change for different tasks, they all follow the same philosophy.
    
\begin{algorithm}
\caption{Pseudo-Label Generation Algorithm}\label{alg} 
\textbf{Input:} Target dataset $X_t$; model $f^{(n-1)}$; decay factor \\ \hspace*{\algorithmicindent} $\lambda_{decay}$. \\
\textbf{Intermediate Result:} $P_{\beta}$, $P_{\alpha}$ are predictions after applying \\ \hspace*{\algorithmicindent} invariance and equivariance transform. \\
 \textbf{Output:} Pseudo-labels $\hat{Y}_{t}^{(n)}$; confidence score $C_{t}^{(n)}$.
\begin{algorithmic}[1]
    \For{$X_{t}^{i}$ in $X_t$} \\ \Comment{\textit{Invariance Consistency}}
        \State $P_{\beta} = f^{(n-1)}(T_{\beta}(X_{t}^{i}))$ \\ \Comment{\textit{Equivariance Consistency}}
        \State $P_{\alpha} = T_{\alpha}^{-1}[f^{(n-1)}(T_{\alpha}(X_{t}^{i}))]$ \\ \Comment{\textit{Self-Ensembling}}
        \State Ensemble $P_{\beta}$ and $P_{\alpha}$ to get ($\hat{Y}_{t}^{(n),i}$, $C_{t}^{(n),i}$) \\ \Comment{\textit{Temporal Consistency}}
            \If{$C_{t}^{(n),i} / C_{t}^{(n),i-1} < \lambda_{decay} $}
                \State $\hat{Y}_{t}^{(n),i} = (\hat{Y}_{t}^{(n),i-1}) + \Delta$
                \State $C_{t}^{(n),i} = \lambda_{decay} * C_{t}^{(n),i-1}$
        \EndIf
    \EndFor
    \State Sort $C_{t}^{(n)}$ and obtain $C_{thresh}$ based on a fixed curriculum learning policy. 
    \State Set $C_{t}^{(n),i} = \mathds{1}(C_{t}^{(n),i} \geq C_{thresh}),~~~\forall i$
\end{algorithmic}
\end{algorithm}
    
\subsection{Pseudo-Label Generation} \label{Pseudo-Label Generation}
    In this section, we explain in details  how to apply these consistencies in practice for generating pseudo-labels and propose the pseudo-label generation method as in Algorithm \ref{alg}. 
    
    
    We address the noisy label problem in two ways. First, we develop an algorithm to generate pseudo-labels using consistency-check to remove false positives, assuming that labels generated using the correct information always satisfy these consistencies. Second, we apply the curriculum learning idea to gradually increase the number of training samples and learn models in an iterative fashion. 
    
     
    For the $n$th iteration, with the previous model $f^{(n-1)}$ obtained from the $(n-1)$th iteration, we iterate through each image $X_{t}^{i}$ in the target dataset $X_t$. $f^{(n-1)}$ is not updated in this process. First, for each image, we apply multiple invariance transform $T_{\beta}$, equivariance transform $T_{\alpha}$ to $X_{t}^{i}$, and ensemble all predictions $P_{\beta}$ and $P_{\alpha}$ to get a pair of estimated labels and confidence scores ($\hat{Y}_{t}^{(n),i}$, $C_{t}^{(n),i}$). 
    
    Second, we use temporal consistency to update weak predictions. For each keypoint, we check whether the current confidence score $C_{t}^{(n),i}$ is strong compared to the one in the previous frame $C_{t}^{(n),i-1}$ with respect to a decay factor $\lambda_{decay}$. If the current frame prediction is confident, we simply keep it; otherwise, we replace the prediction $\hat{Y}_{t}^{(n),i}$ with the flow prediction $\Delta$ plus the previous frame prediction and replace $C_{t}^{(n),i}$ with previous frame confidence multiplied by a decay factor $\lambda_{decay}$. Temporal consistency is optional and can be used if videos are available.
    
    To this end, the algorithm has generated labels and confidence scores for all images. The last step is to iterate through the target dataset again to select $C_{thresh}$ using a curriculum learning strategy, which determines the percentage of labels used for training. The idea here is to use keypoints with high confidence first and gradually include more keypoints after iterations. In practice, we use a policy to include top 20$\%$ ranking keypoints at the beginning, 40$\%$ for the second iteration, until hitting 80$\%$.

\subsection{Consistency-Constrained Semi-Supervised Learning (CC-SSL)}\label{Joint Training}
    For the $n$th iteration, the loss function $L^{(n)}$ is defined to be the Mean Square Error on heatmaps of both the source data and target data, as in Equation \ref{eq:loss}. $\gamma$ is used to balance the loss between source and target datasets.
\begin{align}\label{eq:loss}
\begin{split}
    L^{(n)} =& \sum_i L_{MSE}(f^{(n)}(X_{s}^{i}), Y_{s}^{i}) \\ 
    &+ \gamma \sum_j L_{MSE}(f^{(n)}(X_{t}^{j}), \hat{Y}_{t}^{(n-1),j})
\end{split}
\end{align}

    To this end, we present our Consistency-Constrained Semi-Supervised Learning (CC-SSL) approach as follows: we start with training a model only using synthetic data and obtain an initial weak model $f^{(0)} = f_s$. Then we iterate the following procedure. For the $n$th iteration, we first use Algorithm \ref{alg} to generate labels $\hat{Y}_{t}^{(n)}$. With the generated labels, we simply train the model using ($X_s, Y_s$) and ($X_t, \hat{Y}_{t}^{(n)}$) jointly using $L^{(n)}$.



\subsection{Synthetic Dataset Generation}\label{Synthetic Dataset Generation}
    In order to create a diverse combination of animal appearances and poses, we collect a synthetic animal dataset containing 10+ animals. Each animal comes with several animation sequences. We use Unreal Engine to collect rich ground truth and enable nuisance factor control. The implemented factor control includes randomizing lighting, textures, changing viewpoints and animal poses. 

    The pipeline for generating synthetic data is as follows. Given a CAD model along with a few animation sequences, an animal with random poses and random texture is rendered from a random viewpoint for some random lighting and a random background image. We also generate ground truth depth maps, part segmentation and dense joint locations (both 2D and 3D). See Figure \ref{fig:overview} for samples from the synthetic dataset.

%% file: 04_exp.tex
    \begin{table*}[h]
    \fontsize{8}{10pt}\selectfont
    \setlength{\tabcolsep}{3pt}
    \centering
    \begin{tabular}{l||cccccccc||cccccccc}
    \hline 
     & \multicolumn{8}{c||}{Horse Accuracy} & \multicolumn{8}{c}{Tiger Accuracy}
    \\ 
     & Eye & Chin & Shoulder & Hip & Elbow & Knee & Hoove & Mean & Eye & Chin & Shoulder & Hip & Elbow & Knee & Hoove & Mean\\ [0.5ex] \hline
    \textit{synthetic + real} & \multicolumn{8}{c||}{}\\
    \hspace{0.5cm}Real & 79.04 & 89.71 & \bf 71.38 & 91.78 & 82.85 & 80.80 & 72.76 & 78.98 & \bf 96.77 & 93.68 & 65.90 & 94.99 & 67.64 & 80.25 & 81.72 & 81.99\\
    \hspace{0.5cm}CC-SSL-R & \bf 89.39 & \bf 92.01 & 69.05 & \bf 92.28 & \bf 86.39 & \bf 83.72 & \bf 76.89 & \bf 82.43 & 95.72 & \bf 96.32 & \bf 74.41 & \bf 91.64 & \bf 71.25 & \bf 82.37 & \bf 82.73 & \bf 84.00\\ \hline
    \textit{synthetic only} & \multicolumn{8}{c||}{}\\
    \hspace{0.5cm}Syn & 46.08 & 53.86 & 20.46 & 32.53 & 20.20 & 24.20 & 17.45 & 25.33 & 23.45 & 27.88 & 14.26 & 52.99 & 17.32 & 16.27 & 19.29 & 21.17\\ 
    \hspace{0.5cm}CycleGAN \cite{ DBLP:conf/iccv/ZhuPIE17} & 70.73 & 84.46 & 56.97 & 69.30 & 52.94 & 49.91 & 35.95 & 51.86 & 71.80 & 62.49 & 29.77 & 61.22 & 36.16 & 37.48 & 40.59 & 46.47\\
    \hspace{0.5cm}BDL \cite{DBLP:conf/cvpr/LiYV19} & 74.37 & 86.53 & 64.43 & 75.65 & 63.04 & 60.18 & 51.96 & 62.33 & 77.46 & 65.28 & 36.23 & 62.33 & 35.81 & 45.95 & 54.39 & 52.26\\
    \hspace{0.5cm}CyCADA \cite{DBLP:conf/icml/HoffmanTPZISED18} & 67.57 & 84.77 & 56.92 & 76.75 & 55.47 & 48.72 & 43.08 & 55.57 & 75.17 & 69.64 & 35.04 & 65.41 & 38.40 & 42.89 & 48.90 & 51.48\\ 
    \hspace{0.5cm}CC-SSL & \bf 84.60 & \bf 90.26 & \bf 69.69 & \bf 85.89 & \bf 68.58 & \bf 68.73 & \bf 61.33 & \bf 70.77 & \bf 96.75 & \bf 90.46 & \bf 44.84 & \bf 77.61 & \bf 55.82 & \bf 42.85 & \bf 64.55 & \bf 64.14\\
    \hline

    \end{tabular}
    \caption{Horse and tiger 2D pose estimation accuracy PCK@0.05. Synthetic data are with randomized background and textures. Synthetic only shows results when no real image label is available, Synthetic + Real are cases when real image labels are available. In both scenarios, our proposed CC-SSL based methods achieve the best performance.}
    \label{table:horse-kpts}
    \end{table*}

\section{Experiments}
First, we quantitatively test our approach on the TigDog dataset~\cite{DBLP:conf/cvpr/PeroRSF15} in Section \ref{2D Pose Estimation}. We compare our method with other popular unsupervised domain adaptation methods, such as CycleGAN~\cite{DBLP:conf/iccv/ZhuPIE17}, BDL~\cite{DBLP:conf/cvpr/LiYV19} and CyCADA~\cite{DBLP:conf/icml/HoffmanTPZISED18}. We also qualitatively show keypoint detection of other animals where no labeled real images are available, such as elephants, sheep and dogs. Second, in order to show the domain generalization ability, we annotated the keypoints of animals from Visual Domain Adaptation Challenge dataset (VisDA2019). In Section \ref{Generalization Test on VisDA2019}, we evaluate our models on these images from different visual domains. Third, the rich ground truth in synthetic data enables us to do more tasks beyond 2D pose estimation, so we also visualize part segmentation on horses and tigers and demonstrate the effectiveness of multi-task learning in Section \ref{Part Segmentation}.

\subsection{Experiment Setup}\label{Experiment Setup}
\begin{figure*}
    \centering
    \includegraphics[width=\linewidth, height=5.5cm]{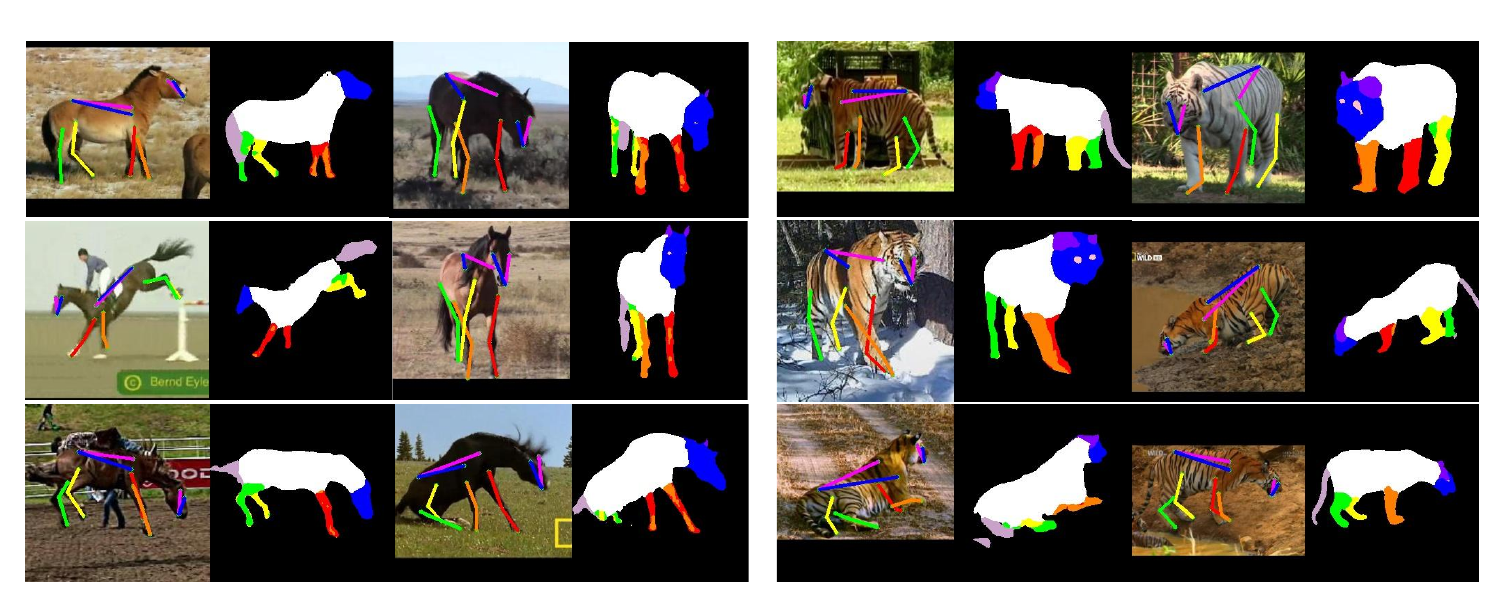}
    \caption{Visualization of horse and tiger 2D pose estimation and part segmentation prediction. The 2D pose estimations are predicted using CC-SSL as described in Section \ref{2D Pose Estimation} and part segmentation predictions are generated using the multi-task learning as described in Section \ref{Part Segmentation}. Best viewed in color.}
    \label{fig:seg-vis}
\end{figure*}

\begin{figure*}
    \centering
    \includegraphics[width=\linewidth,height=5.5cm]{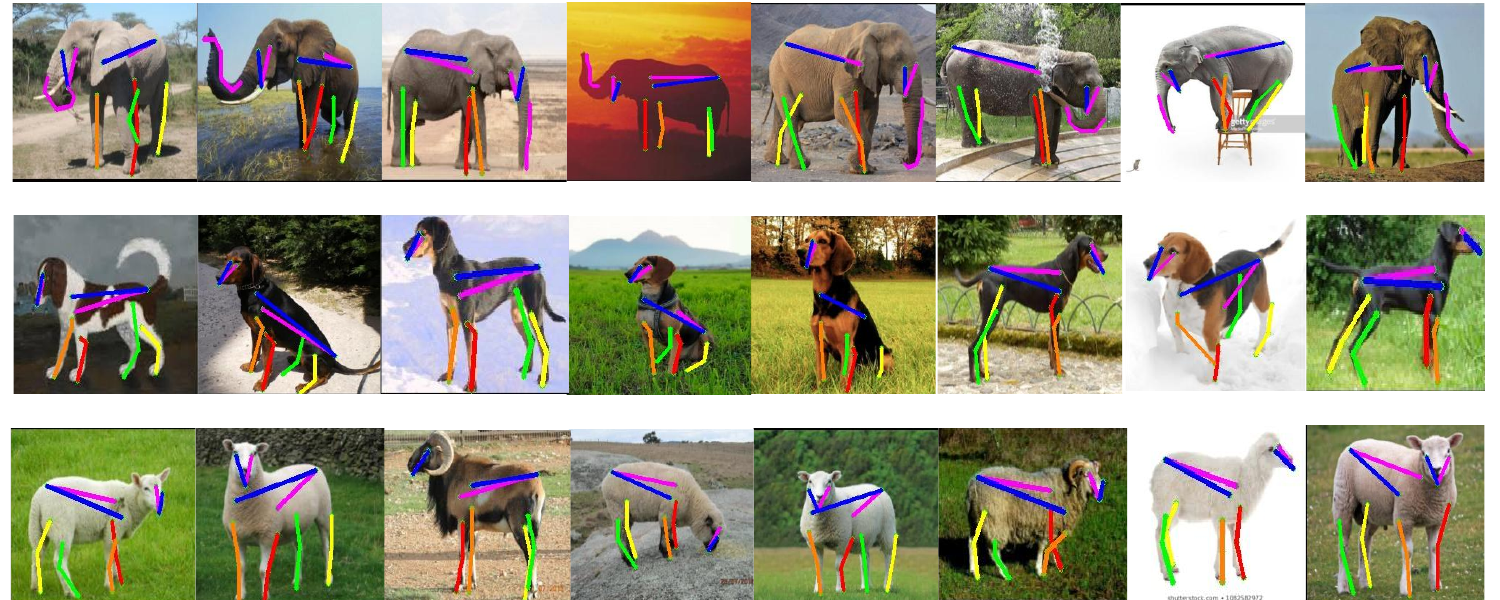}
    \caption{Visualization of 2D pose estimation of other animals. Our method can be easily generalized to elephants' trunks. Best viewed in color.}
    \label{fig:kpts-vis}
\end{figure*}
\textbf{Network Architecture.} We use Stacked Hourglass ~\cite{DBLP:conf/eccv/NewellYD16} as our backbone for all experiments. Architecture design is not our main focus and we strictly follow parameters from the original paper. Each model is trained with RMSProp for $100$ epochs. The learning rate starts with $2.5e^{-4}$ and decays twice at $60$ and $90$ epoches respectively. Input images are cropped with the size of $256\times256$ and augmented with scaling, rotation, flipping and color perturbation.

\textbf{Synthetic Datasets.} We explain the details of our data generation parameters as follows. The virtual camera has a resolution of $640\times480$ and field of view of 90. We randomize synthetic animal textures and backgrounds using Coco val2017 dataset. We does not use any segmentation annotation from coco val2017. For each animal, we generated 5,000 images with random texture and 5,000 images with the texture coming with the CAD model, to which we refer as the original texture. We split the training set and validation set with a ratio of 4:1, resulting in 8,000 images for training and 2,000 for validation. We also generate rich ground truth including part segmentation, depth maps and dense 2D and 3D keypoints. For part segmentation, we define nine parts for each animal, which are eyes, head, ears, torso, left-front leg, left-back leg, right-front leg, right-back leg and tail. The parts definition follows ~\cite{DBLP:conf/cvpr/ChenMLFUY14} with a minor difference that we distinguish front from back legs. CAD models used in this paper are purchased from UE4 marketplace\footnote{\url{https://www.unrealengine.com/marketplace/en-US/product/animal-pack-ultra-01}}.

\textbf{CC-SSL} In our experiments, we pick scaling and rotation from $T_{\alpha}$ and obtain $\Delta$ using optical flow. $\lambda_{decay}$ is set to 0.9 and we train one model for 10 epochs and re-generate pseudo labels with the new model. Models are trained for 60 epochs with $\gamma$ set to be 10.0.

    \textbf{TigDog Dataset} The TigDog dataset is a large dataset containing 79 videos for horses and 96 videos for tigers. In total, for horse, we have 8380 frames for training and 1772 frames for testing. For tigers, we have 6523 frames for training and 1765 frames for testing. Each frame is provided with 19 keypoint annotations, which are defined as eyes(2), chin(1), shoulders(2), legs(12), hip(1) and neck(1). The neck keypoint is not clearly distinguished for left and right, so we leave it out in all experiments. 
    
    
    \subsection{2D Pose Estimation}\label{2D Pose Estimation}
 
    \textbf{Results Analysis.} Our main results are summarized in Table~\ref{table:horse-kpts}. We present our results in two different setups: the first one is under the unsupervised domain adaptation setting where real image annotations are not available; the second one is when labeled real images are available.

    When annotations of real images are not available, our proposed \textbf{CC-SSL} surpasses other methods by a significant margin. The PCK@0.05 accuracy of horses reaches 70.77, which is close to models trained directly on real images. For tigers, the proposed method achieves 64.14. It is worth noticing that these results are achieved without accessing any real image annotation, which demonstrated the effectiveness of our proposed method. 
    
    We also visualize the predicted keypoints in Figure \ref{fig:seg-vis}. Even for some extreme poses, such as horse riding and lying on the ground, the method still generate accurate predictions. The observations for tigers are similar.

    When annotations of real images are available, our proposed \textbf{CC-SSL-R} achieves 82.43 for horses and 84.00 for tigers, which are noticeably better than models trained on real images only. CC-SSL-R is achieved simply by further finetuning the CC-SSL  models using real image labels.

    \begin{table*}[h]
    \fontsize{8}{10pt}\selectfont
    \setlength{\tabcolsep}{3pt}
    \centering
    \begin{tabular}{c||ccc|ccc||ccc|ccc} 
    \hline
    \multirow{3}{*}{} & \multicolumn{6}{c||}{Horse}  & \multicolumn{6}{c}{Tiger}
    \\ \cline{2-13}
     & \multicolumn{3}{c|}{Visible Kpts Accuracy} & \multicolumn{3}{c||}{Full Kpts Accuracy} & \multicolumn{3}{c|}{Visible Kpts Accuracy} & \multicolumn{3}{c}{Full Kpts Accuracy}
    \\ 
      & Sketch & Painting & Clipart &  Sketch & Painting & Clipart & Sketch & Painting & Clipart &  Sketch & Painting & Clipart \\ [0.5ex] \hline 
    Real  & 65.37 & 64.45 & 64.43 & 61.28 & 58.19 & 60.49 & 48.10 & 61.48 & 53.36 & 46.23 &  53.14 & 50.92\\
    CC-SSL & 72.29 & 73.71 & \bf  73.47 & \bf 70.31 & \bf 71.56 & \bf 72.24 & 53.34 & 55.78 & 59.34 & 52.64 & 48.42 & 54.66\\
    CC-SSL-R & \bf 73.25 & \bf 74.56 & 71.78 & 67.82 & 65.15 & 65.87 & \bf 54.94 & \bf 68.12 & \bf 63.47 & \bf 53.43 & \bf 58.66 & \bf 59.29\\
    \hline

    \end{tabular}
    \caption{Horse and tiger 2D pose estimation accuracy PCK@0.05 on VisDA2019. We present our results under two settings: Visible Kpts Accuracy only accounts for visible keypoints; Full Kpts Accuracy also includes self-occluded keypoints. Under all settings, our proposed methods achieve better performance than  baseline Real.}
    \label{table:horse-generalization}
    \end{table*}



    In addition to horses and tigers, we apply the method to other animals as well. The method can be easily transferred to other animal categories and we qualitatively show keypoint prediction results for other animals, as shown in Figure \ref{fig:kpts-vis}. Notice that our method can also detect trunks of elephants.

    We empirically find the performance does not improve much with CycleGAN. We conjecture that one reason is that CycleGAN in general requires a large number of real images to work well. However, in our case, the diversity of real images is limited. Another reason is that animal shapes of transferred images are not maintained well. We also try different adversarial training strategies. Though BDL works quite well for semantic segmentation, we find the improvements on keypoints detection is small. CyCADA also suffers from the same problem as CycleGAN. In comparison, CC-SSL does not suffer from those problems and it can work well even with limited diversity of real data.
    
    We use the same set of augmentations as in \cite{DBLP:conf/eccv/NewellYD16} for baselines \textbf{Real} and \textbf{Syn}. We use a different set of augmentations for other experiments, which we refer to as Strong Augmentation. In addition to what \cite{DBLP:conf/eccv/NewellYD16} used, Strong Augmentation also includes Affine Transform, Gaussian Noise and Gaussian Blurring. 

\subsection{Generalization Test on VisDA2019}\label{Generalization Test on VisDA2019}
    
    In this section, we test model generalization on images from Visual Domain Adaptation Challenge dataset (VisDA2019). The dataset contains six domains: real, sketch, clipart, painting, infograph, and quickdraw. We pick up sketch, painting and clipart for our experiments since inforgraph and quickdraw are not suitable for 2D pose estimation. For each of these three domains, we manually annotate images for horses and tigers. Evaluation results are summarized in Table \ref{table:horse-generalization}. Same as before, we use \textbf{Real} as our baseline. \textbf{CC-SSL} and \textbf{CC-SSL-R} are used for comparison.
    
    For both animals, we observe that models trained using synthetic data achieve best performance in all settings. We present our results under two settings. Visible Keypoints Accuracy only accounts for keypoints that are directly visible whereas Full Keypoints Accuracy shows results with self-occluded keypoints. 
    
    Under all settings, CC-SSL-R is better than Real. More interestingly, notice that even without using real image labels, our CC-SSL method yields better performance than Real in almost all domains. The only one exception is the painting domain of tigers. We hypothesize that this is because texture information (yellow and black stripes) in paintings is still well preserved so models trained on real images can still ``generalize". For sketches and cliparts, appearances are more different from real images and models trained on synthetic data show better results.

\subsection{Part Segmentation} \label{Part Segmentation}

    Since the synthetic animal dataset is generated with rich ground truth, our task is not limited to 2D pose estimation. We also experiment with part segmentation in a multi-task learning setting. All models are trained on synthetic images with Strong Augmentation and tested on TigDog dataset directly. 
    
    As shown in Table~\ref{table:part-segmentation}, we observe that models, trained on keypoints and part segmentation jointly, can generalize better on real images for both animals, compared to the baseline where models are only trained with keypoints. 
    Since we cannot quantitatively evaluate part segmentation predictions, we visualize the part segmentation results on TigDog dataset as shown in Figure~\ref{fig:seg-vis}.
    
    In the multi-task learning setting, we only make minor changes to the original Stacked Hourglass architecture, where we add a branch parallel to the original keypoint prediction one for part segmentation.

    \begin{table}[h]
    \centering
    \begin{tabular}{cccc|cccc} 
    \hline       \hline
    Models & Horse & Tiger
    \\ [0.5ex] \hline 
    Baseline & 60.84 & 50.26\\
    +Part segmentation & \bf 62.25 & \bf 51.69\\
    \hline
    \end{tabular}
    \caption{Horse and tiger 2D pose estimation PCK@0.05 with multi-task learning. We show models can generalize better to real images trained jointly using 2D keypoints and part segmentation.}
    \label{table:part-segmentation}
    \end{table}

%% file: egpaper_for_review.bbl
\begin{thebibliography}{10}\itemsep=-1pt

\bibitem{DBLP:conf/cvpr/AndrilukaPGS14}
Mykhaylo Andriluka, Leonid Pishchulin, Peter~V. Gehler, and Bernt Schiele.
\newblock 2d human pose estimation: New benchmark and state of the art
  analysis.
\newblock In {\em {CVPR}}, pages 3686--3693, 2014.

\bibitem{DBLP:conf/icml/BengioLCW09}
Yoshua Bengio, J{\'{e}}r{\^{o}}me Louradour, Ronan Collobert, and Jason Weston.
\newblock Curriculum learning.
\newblock In {\em {ICML}}, pages 41--48, 2009.

\bibitem{DBLP:journals/corr/abs-1811-05804}
Benjamin Biggs, Thomas Roddick, Andrew~W. Fitzgibbon, and Roberto Cipolla.
\newblock Creatures great and {SMAL:} recovering the shape and motion of
  animals from video.
\newblock {\em CoRR}, abs/1811.05804, 2018.

\bibitem{DBLP:conf/nips/BousmalisTSKE16}
Konstantinos Bousmalis, George Trigeorgis, Nathan Silberman, Dilip Krishnan,
  and Dumitru Erhan.
\newblock Domain separation networks.
\newblock In {\em {NeurIPS}}, pages 343--351, 2016.

\bibitem{DBLP:journals/corr/abs-1908-05806}
Jinkun Cao, Hongyang Tang, Haoshu Fang, Xiaoyong Shen, Cewu Lu, and Yu{-}Wing
  Tai.
\newblock Cross-domain adaptation for animal pose estimation.
\newblock {\em CoRR}, abs/1908.05806, 2019.

\bibitem{DBLP:journals/corr/ChangFGHHLSSSSX15}
Angel~X. Chang, Thomas~A. Funkhouser, Leonidas~J. Guibas, Pat Hanrahan,
  Qi{-}Xing Huang, Zimo Li, Silvio Savarese, Manolis Savva, Shuran Song, Hao
  Su, Jianxiong Xiao, Li Yi, and Fisher Yu.
\newblock Shapenet: An information-rich 3d model repository.
\newblock {\em CoRR}, abs/1512.03012, 2015.

\bibitem{DBLP:conf/3dim/ChenWLSWTLCC16}
Wenzheng Chen, Huan Wang, Yangyan Li, Hao Su, Zhenhua Wang, Changhe Tu, Dani
  Lischinski, Daniel Cohen{-}Or, and Baoquan Chen.
\newblock Synthesizing training images for boosting human 3d pose estimation.
\newblock In {\em {3DV}}, pages 479--488, 2016.

\bibitem{DBLP:conf/cvpr/ChenMLFUY14}
Xianjie Chen, Roozbeh Mottaghi, Xiaobai Liu, Sanja Fidler, Raquel Urtasun, and
  Alan~L. Yuille.
\newblock Detect what you can: Detecting and representing objects using
  holistic models and body parts.
\newblock In {\em {CVPR}}, pages 1979--1986, 2014.

\bibitem{DBLP:journals/corr/abs-1909-00589}
Jaehoon Choi, Taekyung Kim, and Changick Kim.
\newblock Self-ensembling with gan-based data augmentation for domain
  adaptation in semantic segmentation.
\newblock {\em CoRR}, abs/1909.00589, 2019.

\bibitem{DBLP:conf/wacv/DingWFG18}
Yifan Ding, Liqiang Wang, Deliang Fan, and Boqing Gong.
\newblock A semi-supervised two-stage approach to learning from noisy labels.
\newblock In {\em 2018 {IEEE} Winter Conference on Applications of Computer
  Vision, {WACV} 2018, Lake Tahoe, NV, USA, March 12-15, 2018}, pages
  1215--1224, 2018.

\bibitem{DBLP:conf/iccv/DosovitskiyFIHH15}
Alexey Dosovitskiy, Philipp Fischer, Eddy Ilg, Philip H{\"{a}}usser, Caner
  Hazirbas, Vladimir Golkov, Patrick van~der Smagt, Daniel Cremers, and Thomas
  Brox.
\newblock Flownet: Learning optical flow with convolutional networks.
\newblock In {\em {ICCV}}, pages 2758--2766, 2015.

\bibitem{DBLP:conf/iclr/FrenchMF18}
Geoffrey French, Michal Mackiewicz, and Mark~H. Fisher.
\newblock Self-ensembling for visual domain adaptation.
\newblock In {\em {ICLR}}, 2018.

\bibitem{DBLP:conf/eccv/GuoHZZDSH18}
Sheng Guo, Weilin Huang, Haozhi Zhang, Chenfan Zhuang, Dengke Dong, Matthew~R.
  Scott, and Dinglong Huang.
\newblock Curriculumnet: Weakly supervised learning from large-scale web
  images.
\newblock In {\em {ECCV}}, pages 139--154, 2018.

\bibitem{DBLP:conf/icml/HoffmanTPZISED18}
Judy Hoffman, Eric Tzeng, Taesung Park, Jun{-}Yan Zhu, Phillip Isola, Kate
  Saenko, Alexei~A. Efros, and Trevor Darrell.
\newblock Cycada: Cycle-consistent adversarial domain adaptation.
\newblock In {\em {ICML}}, pages 1994--2003, 2018.

\bibitem{DBLP:conf/eccv/HuangLBK18}
Xun Huang, Ming{-}Yu Liu, Serge~J. Belongie, and Jan Kautz.
\newblock Multimodal unsupervised image-to-image translation.
\newblock In {\em {ECCV}}, pages 179--196, 2018.

\bibitem{DBLP:conf/aaai/JiangMZSH15}
Lu Jiang, Deyu Meng, Qian Zhao, Shiguang Shan, and Alexander~G. Hauptmann.
\newblock Self-paced curriculum learning.
\newblock In {\em {AAAI}}, pages 2694--2700, 2015.

\bibitem{DBLP:journals/corr/abs-1908-07387}
Youngdong Kim, Junho Yim, Juseung Yun, and Junmo Kim.
\newblock {NLNL:} negative learning for noisy labels.
\newblock {\em CoRR}, abs/1908.07387, 2019.

\bibitem{DBLP:conf/iclr/LaineA17}
Samuli Laine and Timo Aila.
\newblock Temporal ensembling for semi-supervised learning.
\newblock In {\em {ICLR}}, 2017.

\bibitem{lee2013pseudo}
Dong-Hyun Lee.
\newblock Pseudo-label: The simple and efficient semi-supervised learning
  method for deep neural networks.
\newblock In {\em Workshop on Challenges in Representation Learning, ICML},
  volume~3, page~2, 2013.

\bibitem{DBLP:journals/corr/abs-1906-05586}
Shuyuan Li, Jianguo Li, Weiyao Lin, and Hanlin Tang.
\newblock Amur tiger re-identification in the wild.
\newblock {\em CoRR}, abs/1906.05586, 2019.

\bibitem{DBLP:conf/cvpr/LiYV19}
Yunsheng Li, Lu Yuan, and Nuno Vasconcelos.
\newblock Bidirectional learning for domain adaptation of semantic
  segmentation.
\newblock In {\em {CVPR}}, pages 6936--6945, 2019.

\bibitem{DBLP:conf/nips/LiuBK17}
Ming{-}Yu Liu, Thomas Breuel, and Jan Kautz.
\newblock Unsupervised image-to-image translation networks.
\newblock In {\em {NeurIPS}}, pages 700--708, 2017.

\bibitem{DBLP:conf/icml/LongC0J15}
Mingsheng Long, Yue Cao, Jianmin Wang, and Michael~I. Jordan.
\newblock Learning transferable features with deep adaptation networks.
\newblock In {\em {ICML}}, pages 97--105, 2015.

\bibitem{DBLP:journals/tog/LoperM0PB15}
Matthew Loper, Naureen Mahmood, Javier Romero, Gerard Pons{-}Moll, and
  Michael~J. Black.
\newblock {SMPL:} a skinned multi-person linear model.
\newblock {\em {ACM} Trans. Graph.}, 34(6):248:1--248:16, 2015.

\bibitem{DBLP:conf/cvpr/MurezKKRK18}
Zak Murez, Soheil Kolouri, David~J. Kriegman, Ravi Ramamoorthi, and Kyungnam
  Kim.
\newblock Image to image translation for domain adaptation.
\newblock In {\em {CVPR}}, pages 4500--4509, 2018.

\bibitem{DBLP:conf/eccv/NewellYD16}
Alejandro Newell, Kaiyu Yang, and Jia Deng.
\newblock Stacked hourglass networks for human pose estimation.
\newblock In {\em Computer Vision - {ECCV} 2016 - 14th European Conference,
  Amsterdam, The Netherlands, October 11-14, 2016, Proceedings, Part {VIII}},
  pages 483--499, 2016.

\bibitem{DBLP:conf/bmvc/NovotnyLV16}
David Novotn{\'{y}}, Diane Larlus, and Andrea Vedaldi.
\newblock I have seen enough: Transferring parts across categories.
\newblock In {\em {BMVC}}, 2016.

\bibitem{DBLP:conf/cvpr/PeroRSF15}
Luca~Del Pero, Susanna Ricco, Rahul Sukthankar, and Vittorio Ferrari.
\newblock Articulated motion discovery using pairs of trajectories.
\newblock In {\em {CVPR}}, pages 2151--2160, 2015.

\bibitem{DBLP:conf/icra/PrakashBBACSSB19}
Aayush Prakash, Shaad Boochoon, Mark Brophy, David Acuna, Eric Cameracci,
  Gavriel State, Omer Shapira, and Stan Birchfield.
\newblock Structured domain randomization: Bridging the reality gap by
  context-aware synthetic data.
\newblock In {\em {ICRA}}, pages 7249--7255, 2019.

\bibitem{DBLP:conf/cvpr/RadosavovicDGGH18}
Ilija Radosavovic, Piotr Doll{\'{a}}r, Ross~B. Girshick, Georgia Gkioxari, and
  Kaiming He.
\newblock Data distillation: Towards omni-supervised learning.
\newblock In {\em {CVPR}}, pages 4119--4128, 2018.

\bibitem{DBLP:conf/cvpr/RoyChowdhuryCSJ19}
Aruni {Roy Chowdhury}, Prithvijit Chakrabarty, Ashish Singh, SouYoung Jin,
  Huaizu Jiang, Liangliang Cao, and Erik~G. Learned{-}Miller.
\newblock Automatic adaptation of object detectors to new domains using
  self-training.
\newblock In {\em {CVPR}}, pages 780--790, 2019.

\bibitem{DBLP:conf/cvpr/SappT13}
Benjamin Sapp and Ben Taskar.
\newblock {MODEC:} multimodal decomposable models for human pose estimation.
\newblock In {\em {CVPR}}, pages 3674--3681, 2013.

\bibitem{DBLP:conf/eccv/SunS16}
Baochen Sun and Kate Saenko.
\newblock Deep {CORAL:} correlation alignment for deep domain adaptation.
\newblock In {\em {ECCV} Workshops}, pages 443--450, 2016.

\bibitem{DBLP:conf/cvpr/TremblayPABJATC18}
Jonathan Tremblay, Aayush Prakash, David Acuna, Mark Brophy, Varun Jampani, Cem
  Anil, Thang To, Eric Cameracci, Shaad Boochoon, and Stan Birchfield.
\newblock Training deep networks with synthetic data: Bridging the reality gap
  by domain randomization.
\newblock In {\em {CVPR}}, pages 969--977, 2018.

\bibitem{DBLP:conf/iccv/TzengHDS15}
Eric Tzeng, Judy Hoffman, Trevor Darrell, and Kate Saenko.
\newblock Simultaneous deep transfer across domains and tasks.
\newblock In {\em {ICCV}}, pages 4068--4076, 2015.

\bibitem{DBLP:conf/cvpr/TzengHSD17}
Eric Tzeng, Judy Hoffman, Kate Saenko, and Trevor Darrell.
\newblock Adversarial discriminative domain adaptation.
\newblock In {\em {CVPR}}, pages 2962--2971, 2017.

\bibitem{DBLP:journals/corr/TzengHZSD14}
Eric Tzeng, Judy Hoffman, Ning Zhang, Kate Saenko, and Trevor Darrell.
\newblock Deep domain confusion: Maximizing for domain invariance.
\newblock {\em CoRR}, abs/1412.3474, 2014.

\bibitem{DBLP:conf/cvpr/Varol0MMBLS17}
G{\"{u}}l Varol, Javier Romero, Xavier Martin, Naureen Mahmood, Michael~J.
  Black, Ivan Laptev, and Cordelia Schmid.
\newblock Learning from synthetic humans.
\newblock In {\em {CVPR}}, pages 4627--4635, 2017.

\bibitem{WelinderEtal2010}
P. Welinder, S. Branson, T. Mita, C. Wah, F. Schroff, S. Belongie, and P.
  Perona.
\newblock {Caltech-UCSD Birds 200}.
\newblock Technical Report CNS-TR-2010-001, California Institute of Technology,
  2010.

\bibitem{DBLP:conf/iccv/ZhuPIE17}
Jun{-}Yan Zhu, Taesung Park, Phillip Isola, and Alexei~A. Efros.
\newblock Unpaired image-to-image translation using cycle-consistent
  adversarial networks.
\newblock In {\em {ICCV} 2017}, pages 2242--2251, 2017.

\bibitem{DBLP:conf/eccv/ZouYKW18}
Yang Zou, Zhiding Yu, B.~V. K.~Vijaya Kumar, and Jinsong Wang.
\newblock Unsupervised domain adaptation for semantic segmentation via
  class-balanced self-training.
\newblock In {\em {ECCV}}, pages 297--313, 2018.

\bibitem{DBLP:journals/corr/abs-1908-09822}
Yang Zou, Zhiding Yu, Xiaofeng Liu, B.~V. K.~Vijaya Kumar, and Jinsong Wang.
\newblock Confidence regularized self-training.
\newblock {\em CoRR}, abs/1908.09822, 2019.

\bibitem{DBLP:journals/corr/abs-1908-07201}
Silvia Zuffi, Angjoo Kanazawa, Tanya~Y. Berger{-}Wolf, and Michael~J. Black.
\newblock Three-d safari: Learning to estimate zebra pose, shape, and texture
  from images "in the wild".
\newblock {\em CoRR}, abs/1908.07201, 2019.

\bibitem{DBLP:conf/cvpr/ZuffiKB18}
Silvia Zuffi, Angjoo Kanazawa, and Michael~J. Black.
\newblock Lions and tigers and bears: Capturing non-rigid, 3d, articulated
  shape from images.
\newblock In {\em {CVPR}}, pages 3955--3963, 2018.

\bibitem{DBLP:conf/cvpr/ZuffiKJB17}
Silvia Zuffi, Angjoo Kanazawa, David~W. Jacobs, and Michael~J. Black.
\newblock 3d menagerie: Modeling the 3d shape and pose of animals.
\newblock In {\em {CVPR}}, pages 5524--5532, 2017.

\end{thebibliography}
